\documentclass{optica-article}

\journal{opticajournal} 

\articletype{Research Article}

\usepackage{subcaption}   
\usepackage{float}
\usepackage{graphicx}
\usepackage{subcaption}
\usepackage{caption}
\usepackage{xcolor}
\usepackage{soul}

\usepackage{booktabs}

\begin{document}
\title{Clinical Feasibility of Low-Magnification Fluorescence Imaging for Breast Cancer Margin Detection Using Texture Analysis and Deep Learning}

\author{Tianling Niu \authormark{1+},
Pouya Afshin \authormark{2+},
Tongtong Lu \authormark{3},
David Helminiak \authormark{4},
Julie Jorns \authormark{5},
Mollie Patton \authormark{5},
Tina Yen \authormark{6},
Dong Hye Ye \authormark{2},
and Bing Yu \authormark{1*}
}

\address{\authormark{1}Joint Department of Biomedical Engineering, Marquette University and Medical College of Wisconsin, Milwaukee, WI, USA\\
\authormark{2}Department of Computer Science, Georgia State University, Atlanta, GA, USA\\
\authormark{3}Department of Engineering and Engineering Technology, University of Wisconsin--Oshkosh, Oshkosh, WI, USA\\
\authormark{4}Department of Computer Engineering, Marquette University, Wisconsin, WI, USA\\
\authormark{5}Department of Pathology, Medical College of Wisconsin, Milwaukee, WI, USA\\
\authormark{6}Department of Surgery, Medical College of Wisconsin, Milwaukee, WI, USA}
\email{\authormark{*}Corresponding author: bingyu@marquette.edu}

\noindent\authormark{+}These authors contributed equally to this work.


\begin{abstract*}

High-resolution images of unprocessed surgical breast tissue can be obtained using microscopy with ultraviolet surface excitation (MUSE). This technique is considered a promising method for checking surgical margins during breast cancer surgery. In this study, MUSE images at 4$\times$ and 10$\times$ magnifications were compared using patch-level classification methods. Texture analysis (TA) based on local binary patterns (LBP) and deep learning (DL) with a base Vision Transformer (ViT) model were used. Both methods achieved similar performance at both magnifications. Using DL method, both 4$\times$ and 10$\times$ magnifications achieved 96.30\% sensitivity, 100\% specificity and 98.18\% accuracy. Using TA method, 4$\times$ achieved better specificity (100\% vs 93.33\%) and 10$\times$ yielded higher sensitivity (100\% vs 93.33\%), but both had the same accuracy (96.67\%). No clear improvement in performance was observed with 10× magnification. These results show that 4$\times$ imaging achieves the same diagnostic accuracy as 10$\times$ imaging. At the same time, 4$\times$ offers a larger field of view and faster image capture. Therefore, lower magnification can be effectively used in MUSE systems for accurate and efficient intraoperative margin assessment.
\end{abstract*}

\section{Introduction}
Breast cancer has had the highest incidence among cancers. It is the second leading cause of cancer-related death among women in the United States, even with improvements in screening and treatment \cite{Siegel2025, Davis2025}. Surgical treatment is the most prevalent option for breast cancer patients, and breast-conserving surgery is recommended for the earliest-stage patients \cite{Czajka2023}. However, approximately 20\% of patients require secondary surgery due to incomplete excision during the primary procedure \cite{Kaczmarski2019}. This indicates that achieving complete tumor removal remains a challenge in breast-conserving surgery. The surgical margin, the portion of the patient's removed tissue, is examined using hematoxylin and eosin (H\&E) staining. This process usually takes several days because the tissue must first be fixed in formalin and embedded in paraffin (FFPE) before analysis \cite{Tseng2023}. Margins are classified as either negative or positive; the latter indicates that cancer cells remain at the surgical margin, thereby increasing the risk of recurrence. Therefore, additional surgery is often required to obtain a negative margin \cite{Nayyar2018}. Secondary surgery can place both emotional and financial burdens on patients, highlighting the need for methods that allow real-time, intraoperative margin assessment during breast-conserving surgery \cite{Mullenix2004}.

Microscopy with ultraviolet surface excitation (MUSE) applies ultraviolet (UV) light to image the tissue surface at high resolution rapidly and without sectioning, making it a promising approach for real-time margin evaluation \cite{Fereidouni2017, Xie2019, Lu2020, Niemeier2022, Yoshitake2018}. Unlike traditional H\&E, which requires thin tissue sections and relies heavily on pathologist expertise, MUSE offers rapid surface imaging with distinctive contrast patterns that enable clinicians to distinguish healthy from cancerous tissue and to clearly identify tumor boundaries, both of which are essential during surgery. The shallow penetration of UV light limits imaging to the tissue surface without the need for sectioning, which greatly shortens image acquisition time and makes MUSE suitable for intraoperative assessment \cite{Fereidouni2017, Xie2019, Lu2020, Niemeier2022, Yoshitake2018}. Moreover, the MUSE system has a simple optical setup and does not require complex components, making it a cost-effective method for margin assessment that is easy to use in clinical settings \cite{Fereidouni2015}.  


Beyond intraoperative margin assessment, MUSE has been applied to rapid histology, surface topography, and nerve imaging \cite{Xie2019, Lu2020, Niemeier2022, Yoshitake2018, Fereidouni2015, Kolluru2022, Levenson2016}. Its objective magnification can be adjusted: higher magnifications (10$\times$ or 20$\times$) capture finer details when speed is less critical. This allows MUSE to complement conventional histopathology and visualize nerve microanatomy \cite{Xie2019, Lu2020, Niemeier2022, Yoshitake2018, Fereidouni2015, Kolluru2022, Levenson2016}. For example, Levenson et al. used 10$\times$ for surface tomography and histopathology \cite{Levenson2016, Fereidouni2015, Xie2019}, while others used 10$\times$ for nerve morphology \cite{Kolluru2022} and breast margins \cite{Lu2020}, or 10–20$\times$ for skin and breast tissue imaging \cite{Yoshitake2018}. Higher magnification reveals finer details but captures a smaller field of view (FOV), slowing scanning of the entire tissue, an important factor for rapid imaging during surgery. To address this limitation, a MUSE system with a 4$\times$ magnification was developed, achieving classification accuracies of 90.3\% with texture analysis (TA) and 96.47\% with deep learning (DL) \cite{Lu2022, Afshin2026}, demonstrating that computational approaches can compensate for reduced visual detail.

Building on this, TA provides a simple, computationally efficient method for quantifying tissue patterns. Local binary pattern (LBP), a common TA method, compares each pixel with its neighbors and encodes these relationships to capture micro-texture. LBP performs well for classifying MUSE fluorescence breast images with minimal computational cost \cite{Song2013, Lu2022, Heikkila2009}. Although TA is fast and straightforward, DL can automatically capture more complex patterns from image patches, including fine cellular details and overall tissue structure. When applied to MUSE images, DL improves margin assessment, reduces the need for follow-up surgeries, and supports more accurate detection and treatment decisions for breast cancer \cite{Afshin2025, Afshin2026}.

Because MUSE images are very large, analyzing them at high resolution with DL is computationally challenging, even at 4$\times$ magnification. Downsampling can also remove fine details needed for accurate classification. To solve this, researchers divide MUSE images into smaller patches, analyze each patch separately, and then combine the results to classify tissue \cite{To2023, Lu2020, Lu2022}. Many studies have used Convolutional Neural Networks (CNNs) \cite{LeCun1989}, such as ResNet-50 with XGBoost \cite{To2023}, or have improved training with synthetic patches generated by diffusion models \cite{Ghahfarokhi2024}. CNNs work well but may miss overall tissue context due to the limited field of view \cite{Afshin2025}. Vision transformers (ViTs) capture both fine details and the full tissue structure, using self-attention to capture relationships among subpatches within each patch. Patch-level ViTs, when combined with Grad-CAM++, can highlight important regions and facilitate interpretation of predictions \cite{Afshin2025}. In contrast, others used self-supervised synthetic MUSE images to enhance training and classification accuracy further \cite{Afshin2026}. Inspired by these results, the current study uses a patch-level ViT framework to analyze MUSE images.

\begin{figure*}[htbp]
    \centering
    \includegraphics[width=0.95\linewidth]{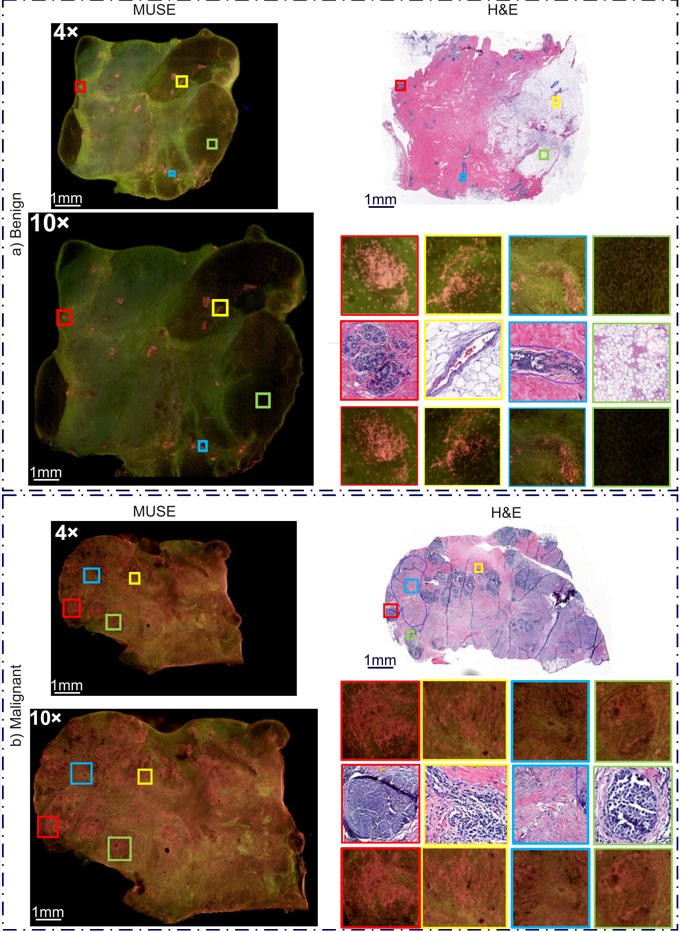}
   \caption{
Representative MUSE and H\&E images of (a) benign tissue and (b) invasive lobular carcinoma (ILC). 
For each case, 4$\times$ MUSE (top left), H\&E (top right), and 10$\times$ MUSE (bottom left) images are shown, with corresponding zoomed-in regions (bottom right) that highlight key morphological structures between tissue types. 
In (a), glands (red), vessels (yellow), ducts (blue), and adipocytes (green) are indicated; in (b), lobular neoplasia (red/green), chronic inflammation (yellow), and high tumor cellularity (blue) are highlighted.
}
    \label{fig:b-m}
\end{figure*}
From an optical perspective, higher magnification provides better resolution but also increases acquisition time. For example, acquiring an image of all scanned breast tissues generally takes a few minutes with a 4$\times$ objective but more than 15 minutes with a 10$\times$ objective. Additionally, higher-magnification objectives have a shallower depth of field and a narrower field of view, which require more precise focusing and longer exposure times. Therefore, there is a clear trade-off between image resolution and imaging speed in the evaluation of intraoperative margins \cite{Niu2025}. Although previous studies have reported promising results with MUSE in various applications, most have evaluated it at a single magnification. As a result, it remains unclear how different magnifications affect classification performance, particularly for binary classification of tumor versus normal tissue \cite{Fereidouni2017, Xie2019, Lu2020, Niemeier2022, Yoshitake2018, Fereidouni2015, Kolluru2022, Levenson2016}. To address this gap, this study compares the classification performance of 4$\times$ and 10$\times$ magnifications using texture analysis and deep learning models \cite{Lu2022, To2023, Afshin2025}. By directly comparing them, we provide clear results to inform the selection of the most suitable objective for intraoperative assessment with MUSE. In particular, this study investigates whether 10$\times$  yields a significant improvement in classification accuracy or whether 4$\times$  suffices while allowing faster image acquisition.

\begin{figure}[t]
\centering
\includegraphics[width=\linewidth]{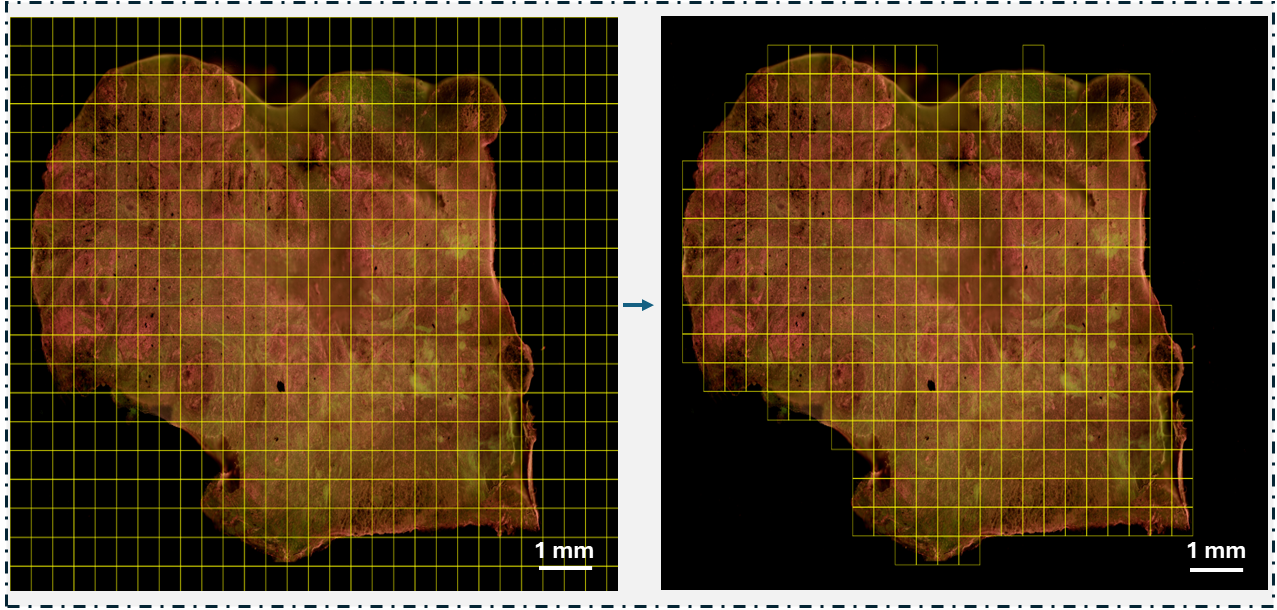}
\caption{Illustration of patch extraction from a 10$\times$ MUSE whole-slide image of a grade 2 invasive lobular carcinoma (ILC) sample. Patches are selected using a 20\% threshold, retaining only those with more than 20\% of pixels above intensity 5 (background). Extracted patches are square, sized $1000 \times 1000$ pixels for 10$\times$.}
\label{fig:patch_extraction}
\end{figure}

\section{Materials and Methods}

\subsection{Pre-processing}
Before classification, all fluorescence images are prepared to ensure consistent and informative input data. The fluorescence dyes used in this study are propidium iodide (PPID) and eosin Y (EY). PPID stains the nuclei and produces red fluorescence, while EY stains the cytoplasm and produces green fluorescence. All fluorescence images are stitched MUSE images, as shown in Fig.~\ref{fig:b-m} a) presents a representative example of benign breast tissue, including MUSE images acquired at 4$\times$ and 10$\times$ magnifications, along with the corresponding H\&E images for comparison. Zoomed-in views are provided to highlight benign glands, blood vessels, benign ducts, and adipocytes, each indicated by a distinct color. Fig.~\ref{fig:b-m}  b) shows a representative example of malignant breast tissue, including MUSE images acquired at 4$\times$ and 10$\times$ magnifications, along with the corresponding H\&E images for comparison. Zoomed-in regions highlight lobular neoplasia, chronic inflammation, and areas of high cellular density. The fluorescence images were divided into patches for analysis because the original stitched images were very large, often exceeding 10,000 $\times$ 10,000 pixels.

During the preprocessing, the stitched MUSE images \( \{x_i\}_{i=1}^{M} \in \mathcal{X}\) are divided into non-overlapping patches \(\mathbf{p}_i^j\) of size 400$\times$400 pixels for 4$\times$ magnification and 1000$\times$1000 pixels for 10$\times$ magnification. The patch sizes were scaled according to the magnification difference (10$\times$ / 4$\times$ = 2.5) to maintain an approximately consistent physical tissue area across different imaging magnifications. The patch extraction process is indicated in Fig.~\ref{fig:patch_extraction}. The red channel is extracted, and a threshold of 5 is applied to identify informative pixels, based on empirical observations. Pixels with intensity above the threshold are considered informative, and only patches \(\mathbf{p}_i^j\) containing more than 20\% informative pixels are retained for classification; all other patches are discarded. 

Patch labels were generated based on the corresponding H\&E images using semi-automated annotation transfer method and verified by our pathologist\cite{niu2026effective}. Patches from benign regions were labeled as normal, while patches from malignant regions were labeled as tumor.  To minimize the risk of excluding tumor tissue due to boundary uncertainty, patches extracted from the transitional borders between benign and malignant zones were labeled as tumor. Let \(J_i = \{ j \mid \mathbf{p}_i^j \in x_i \}\) denote the set of indices of all valid patches extracted from MUSE image \(x_i\). Each patch \(\mathbf{p}_i^j\)  is labeled according to the corresponding image of H\&E, which is the gold standard clinically. Patches along the edge of cancerous tissue were also labeled as malignant.

 In this study, two approaches are used for margin classification: a traditional TA method and a DL method. In the TA approach, texture features are extracted from image patches and are used for classification with a conventional machine learning model. In the DL approach, discriminative features are automatically learned from image patches using a fine-tuned Vision Transformer model. In both methods, patch-level predictions are aggregated to obtain the final margin-level classification.



\subsection{Texture Analysis}

Texture analysis has been an effective method for capturing local patterns and structural information in medical images, helping to differentiate normal from cancerous tissue based on textural differences \cite{Wan2017}. In this study, texture features from the red channel of each MUSE patch \(\mathbf{p}_i^j\) are extracted using the local binary pattern (LBP) method \cite{Lu2022, Kral2016, Pietikainen2010, Liao2009}. Red channel is used because it represents the nuclei information and provides the highest contrast. Before extracting LBP features, the patches are denoised using a Wiener filter and histogram equalized to ensure uniform pixel intensities. These features are then used for classification. Specifically, from each patch \(\mathbf{p}_i^j\), \(j \in \mathcal{J}_i\), 14 LBP features are extracted using the number of neighbors \(N = 12\) and a circular pattern radius \(R = 3\). This configuration samples 12 evenly spaced points on a circle of radius 3 centered at each pixel and compares the intensity values at these locations with that of the center pixel to encode local texture patterns.

\begin{figure}[t]
\centering
\includegraphics[width=\linewidth]{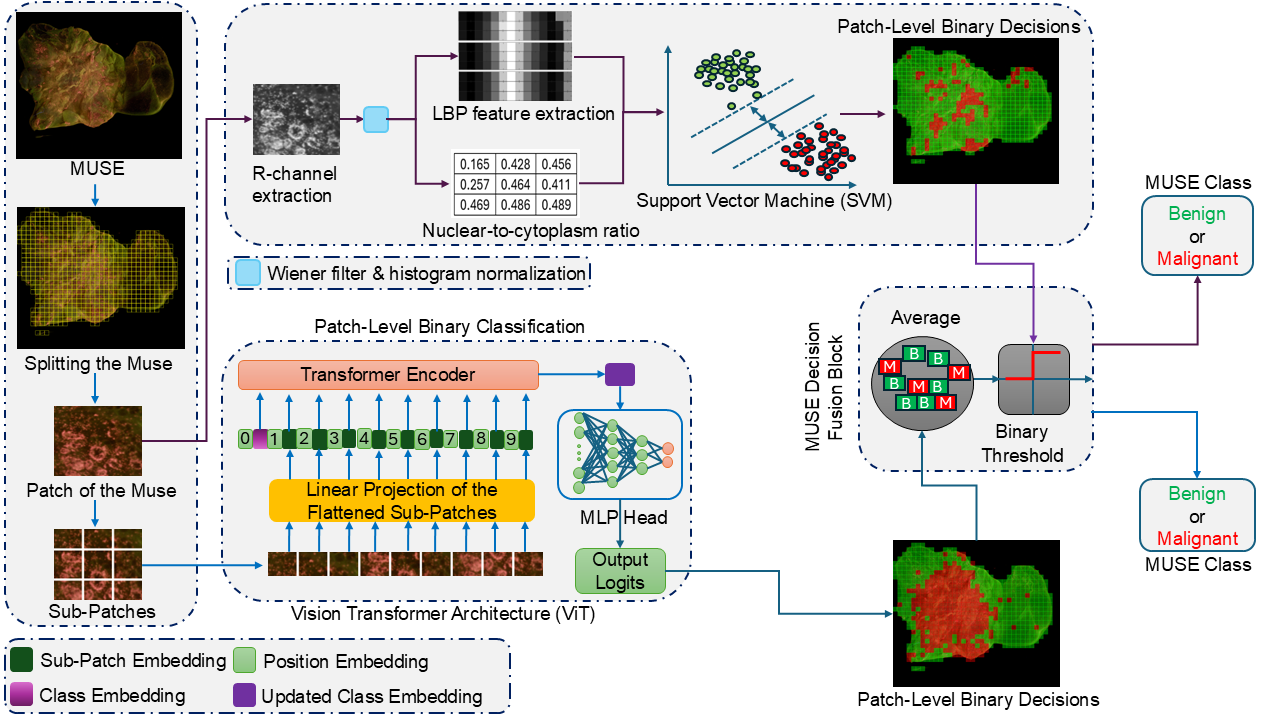} 
\caption{Workflow for classifying MUSE images using TA (top branch) and DL (bottom branch). MUSE images are divided into patches at two magnifications. For TA, the red channel of each patch is first preprocessed. Then, LBP features are extracted and an SVM model predicts the class of each patch. For DL, each patch is converted into embeddings and passed through a ViT model with an MLP head to generate patch-level predictions. In both methods, the patch predictions are combined and compared with a threshold to determine the final margin-level classification.}
\label{system}
\end{figure}

Linear interpolation and rotation-invariant uniform patterns are used in feature computation to ensure robust LBP feature extraction when the sampled neighbor points do not align exactly with pixel grid locations and to reduce sensitivity to image rotation. The extracted LBP features are stored in a feature matrix \(\mathbf{F}_i \in \mathbb{R}^{J_i \times 14}\). The nuclei-to-cytoplasm (NC) ratio from each patch \(\mathbf{p}_i^j\) is also calculated using a dark threshold of 115 (set empirically) and stored in a separate matrix \(\mathbf{R}_i \in \mathbb{R}^{J_i \times 1}\). The feature matrices \(\mathbf{F}_i\) and \(\mathbf{R}_i\), together with the healthy/malignant labels \(y_i^j\) for \(j \in \mathcal{J}_i\), are used as inputs to a support vector machine (SVM) to perform classification and predict labels for the test set, which works by drawing a boundary that best separates different tissue types based on extracted image features. The polynomial kernel helps the model draw a curved boundary rather than a straight line, which is useful when the data are not easily separable. The regularization parameter \( C \) was set to 20 in MATLAB and controls how strictly the boundary follows the training data: a higher value places more emphasis on correctly classifying the training samples, while still limiting overly complex boundaries. 

Then five-fold cross-validation is used to evaluate patch-level classification performance. Min-max normalization is applied to all features to prevent data leakage prior to margin-level classification. The trained patch classifier produces class predictions $\hat{y}_i^j$ for all valid patches $\mathbf{p}_i^j \in J_i$. The tumor and normal predictive scores for the MUSE image $x_i$ were computed as

\begin{align}
S_i^{\text{tumor}} = \sum_{j \in \mathcal{J}_i} \hat{y}_i^j,
\end{align}
and the normal predictive score was computed as
\begin{align}
S_i^{\text{normal}} = \sum_{j \in \mathcal{J}_i} (1 - \hat{y}_i^j).
\end{align}

The final margin-level label \(\hat{y}_i\) was determined by comparing the tumor and normal predictive scores:
\begin{align}
\hat{y}_i =
\begin{cases}
1, & \text{if } S_i^{\text{tumor}} > \theta  \times S_i^{\text{normal}},\\
0, & \text{otherwise}.
\end{cases}
\end{align}

A margin was classified as malignant when the tumor prediction score exceeded the normal prediction score scaled by a threshold; otherwise, it was classified as benign. The complete workflow is illustrated in Fig.~\ref{system}.

\subsection{Deep Learning with Patch-Level ViT}

A pre-trained ViT \cite{Dosovitskiy2020} is used to classify patches from MUSE Images and aggregate the results for margin-level predictions. Using transfer learning, the model recognizes common image features, thereby reducing overfitting and accelerating training \cite{Weiss2016}. After fine-tuning the MUSE patch-level dataset, the model extracts detailed features and classifies each patch as benign or malignant. To that end,
each extracted patch $\mathbf{p}_i^j$ is divided into $N$ non-overlapping sub-patches 
$\mathbf{s}_i^{jk}$ (see Fig.~\ref{system}), which are reshaped and linearly projected into a $D$-dimensional 
embedding space $\hat{\mathbf{s}}_i^{jk} = \mathrm{vec}(\mathbf{s}_i^{jk}) \mathbf{E} 
\in \mathbb{R}^{1 \times D}$ using a trainable projection matrix 
$\mathbf{E} \in \mathbb{R}^{(P^2 \cdot C) \times D}$.
A learnable class token $\mathbf{p}_i^{j\text{class}} \in \mathbb{R}^{1 \times D}$ is 
prepended to represent global patch information, and positional embeddings 
$\mathbf{E}_i^{j\text{pos}} \in \mathbb{R}^{(N+1) \times D}$ are added to encode spatial 
relationships. The resulting input sequence is given by \cite{Afshin2026}
\begin{equation}
\mathbf{h}_0 =
\big[ \mathbf{p}_i^{j\text{class}}, 
\hat{\mathbf{s}}_i^{j1}, \dots, \hat{\mathbf{s}}_i^{jN} \big]
+ \mathbf{E}_i^{j\text{pos}} .
\end{equation}

The sequence $\mathbf{h}_0$ is processed by a vision transformer encoder composed of $L$ 
layers with normalization, self-attention, and multi-layer perceptron (MLP) blocks. 
The final class token is normalized and passed to a classification head to predict the 
patch label \cite{Afshin2026}:
\begin{equation}
\hat{y}_i^j =
\mathrm{MLP}_{\mathrm{head}}
\big( \mathrm{LN}(\mathbf{p}_i^{j\text{class}}) \big)
\in \{0,1\},
\end{equation}
where $0$ and $1$ denote benign and malignant classes, respectively.
The margin-level label \(\hat{y}_i\) is obtained by averaging the patch
predictions over the valid patch indices \(\mathcal{J}_i\) and comparing
to a threshold \(\theta\) \cite{Afshin2026}:
\[
\hat{y}_i =
\begin{cases}
1, & \text{if } \frac{1}{|\mathcal{J}_i|}
\sum_{j \in \mathcal{J}_i} \hat{y}_i^j > \theta,\\
0, & \text{otherwise}.
\end{cases}
\]
For the $4\times$ and $10\times$ system setups, the $4\times$ model was optimized with the AdamW optimizer at a learning rate of $5 \times 10^{-5}$ and a weight decay of $1 \times 10^{-4}$. In contrast, for the 10$\times$  setting, a higher learning rate of $4 \times 10^{-4}$ was employed. For the $10\times$ configuration, the model was trained with stochastic gradient descent (SGD) with a momentum of 0.9 and the same weight decay parameter. Preliminary ablation experiments using weighted cross-entropy loss to address patch imbalance did not improve performance and yielded no consistent gains in sensitivity or specificity; therefore, the final models were trained with the standard unweighted loss formulation. All models were trained for 60 epochs. Patch-level classification was performed using the ViT-B/16 architecture with 5-fold cross-validation. The data split was performed at the margin-level to prevent data leakage. In each fold, a set of MUSE images was held for testing, while patches extracted from the remaining MUSE images were further divided into 80\% for training and 20\% for validation.
 \begin{table*}[ht]
\centering
\caption{Dataset summary at 4$\times$ and 10$\times$ magnifications including margin-level and patch-level distributions.}
\label{tab:sample_patch}
\resizebox{\textwidth}{!}{%
\begin{tabular}{l c c c c | c c c c}
\hline
& \multicolumn{4}{c|}{4$\times$ magnification} 
& \multicolumn{4}{c}{10$\times$ magnification} \\
\cline{2-9}
& Samples & Margin (\%) & Patches & Patch (\%) 
& Samples & Margin (\%) & Patches & Patch (\%) \\
\hline
Healthy   & 28 & 50.9 & 29,897 & 76.4 
          & 28 & 50.9 & 27,419 & 75.1 \\
Malignant & 27 & 49.1 & 9,248  & 23.6 
          & 27 & 49.1 & 9,078  & 24.9 \\
\hline
Total     & 55 & 100 & 39,145 & 100 
          & 55 & 100 & 36,497 & 100 \\
\hline
\end{tabular}}
\end{table*}

\begin{figure*}[t]
    \centering
    \begin{subfigure}[b]{0.497\linewidth}
        \centering
        \includegraphics[width=\linewidth]{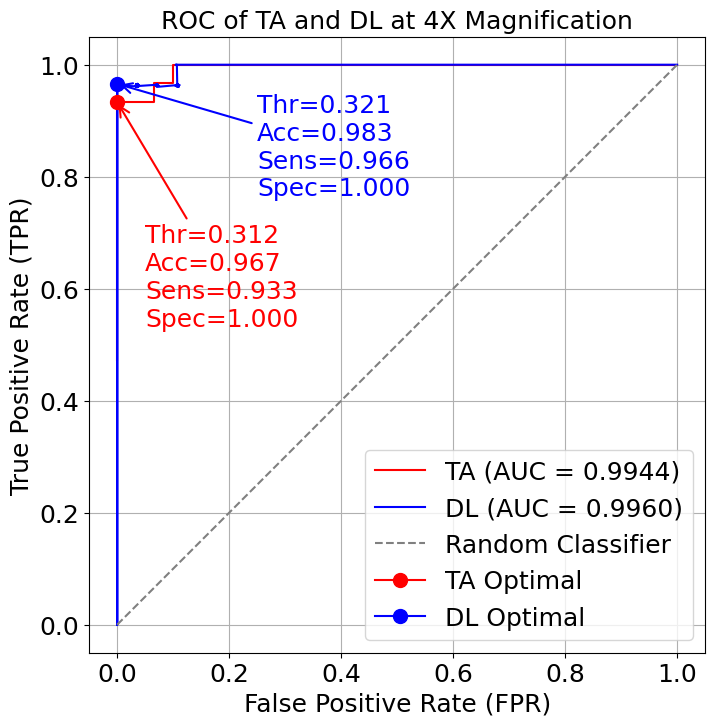}
        \caption{ROC Curve for DL vs TA (4$\times$)}
        \label{fig:roc_4x}
    \end{subfigure}
    \hfill
    \begin{subfigure}[b]{0.497\linewidth}
        \centering
        \includegraphics[width=\linewidth]{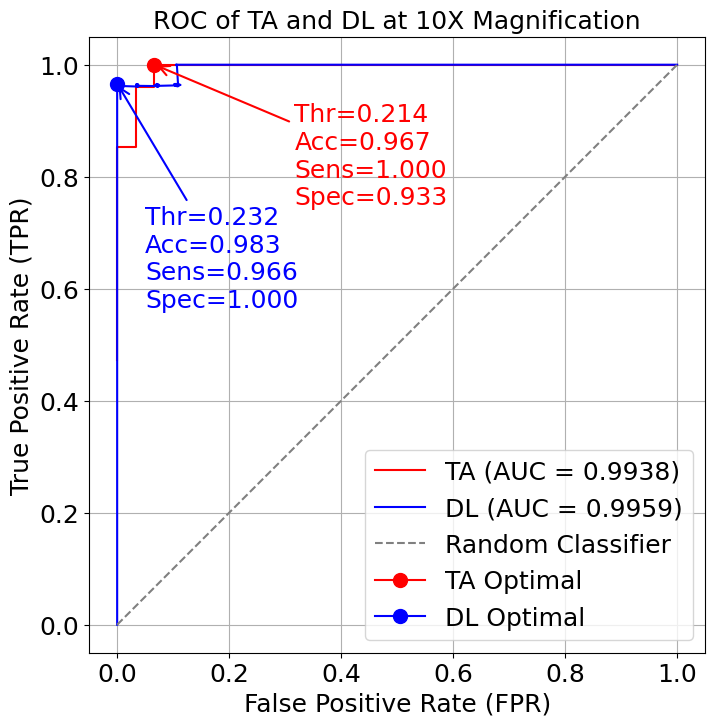}
        \caption{ROC Curve for DL vs TA (10$\times$)}
        \label{fig:roc_10x}
    \end{subfigure}
    \caption{ROC curves comparing margin-level classification performance of DL and TA at both magnifications.}

    \label{fig:roc}
\end{figure*}





\section{Results}
\subsection{Dataset and Experimental Setup}
The dataset in this study is obtained from the Medical College of Wisconsin. It consists of 55 breast tissue samples, including 28 benign and 27 malignant cases. From these samples, a total of 39,145 patches are extracted at a magnification of 4$\times$ and 36,497 patches at a magnification of 10$\times$. At 4$\times$, 29,897 patches are from benign tissue and 9,248 patches from malignant tissue, whereas at 10$\times$, there are 27,419 benign patches and 9,078 malignant patches. Although the margin-level dataset is nearly balanced (50.9\% benign vs. 49.1\% malignant), the patch-level dataset is imbalanced across both magnifications, with a higher proportion of benign patches. This imbalance reflects the larger spatial extent of benign tissue within surgical margins. A summary of the patch-level and margin-level datasets is presented in Table~\ref{tab:sample_patch}.


\subsection{Quantitave Results}
To evaluate classification performance fairly, the best binary threshold was selected for each magnification using grid search along the ROC curves as shown in Figure~\ref{fig:roc}. Instead of using a random cutoff value, all possible thresholds were tested, and the threshold that gave the best margin-level performance was chosen. This ensures that both TA and DL were evaluated under their best operating conditions and allows a reliable comparison between 4$\times$ and 10$\times$ magnifications.
To compare the two methods, both TA and DL were evaluated using the same margin-level classification framework.

Table~\ref{tab:margin_results_combined} summarizes the margin-level results at 4$\times$ and 10$\times$ magnifications. At 4$\times$, TA achieved 96.67\% accuracy, 93.33\% sensitivity, and 100.00\% specificity at the optimal threshold of 0.312 (see Fig. \ref{fig:roc_4x}). Two malignant margins were classified as negative.
At 10$\times$, the accuracy remained 96.67\%. Sensitivity increased to 100.00\%, while specificity decreased to 93.33\% at the optimal threshold of 0.214  (see Fig.\ref{fig:roc_10x}), resulting in two false positives. Although different errors were observed, the overall accuracy was the same at both magnifications. This means that using 10$\times$ does not significantly improve LBP-based margin classification. With an appropriate threshold, 4$\times$ magnification already provides sufficient texture information for reliable classification. In addition, 4$\times$ imaging gives a larger field of view and faster image capture. This makes it more practical and efficient for real-time margin assessment without reducing performance.




\begin{table}[t]
\centering
\caption{Margin-level classification results for TA and DL at 4$\times$ and 10$\times$ magnifications.}
\label{tab:margin_results_combined}
\resizebox{\textwidth}{!}{%
\begin{tabular}{l|l c c c c c|c c c c c}
\toprule
\multirow{2}{*}{Method} & \multirow{2}{*}{} 
& \multicolumn{5}{c|}{4$\times$} 
& \multicolumn{5}{c}{10$\times$} \\
\cmidrule(r){3-7} \cmidrule(r){8-12}
 & & P & N & Total & Se (\%) & Sp (\%) 
 & P & N & Total & Se (\%) & Sp (\%) \\
\midrule

\multirow{3}{*}{TA}
 & P     & 25 & 0  & 25 & 93.33 & --
 & 27 & 2  & 29 & 100.00 & -- \\
 & N     & 2  & 28 & 30 & -- & 100.00
 & 0  & 26 & 26 & -- & 93.33 \\
 & Total & 27 & 28 & 55 & -- & --
 & 27 & 28 & 55 & -- & -- \\
\cmidrule(l){2-12}
 & Accuracy (\%)
 & \multicolumn{5}{c|}{96.67}
 & \multicolumn{5}{c}{96.67} \\

\specialrule{1.5pt}{2pt}{2pt}

\multirow{3}{*}{DL}
 & P     & 26 & 0  & 27 & 96.30 & --
 & 26 & 0  & 27 & 96.30 & -- \\
 & N     & 1  & 28 & 29 & -- & 100.00
 & 1  & 28 & 29 & -- & 100.00 \\
 & Total & 27 & 28 & 55 & -- & --
 & 27 & 28 & 55 & -- & -- \\
\cmidrule(l){2-12}
 & Accuracy (\%)
 & \multicolumn{5}{c|}{98.18}
 & \multicolumn{5}{c}{98.18} \\

\bottomrule
\end{tabular}%
}
\end{table}


The DL framework also showed consistent margin-level performance at both magnifications, as shown in Table~\ref{tab:margin_results_combined}. At 4$\times$ magnification, the model achieved 98.18\% accuracy, 96.30\% sensitivity, and 100.00\% specificity at the optimal threshold of 0.321 (see Fig. \ref{fig:roc_4x}). It correctly identified 26 of 27 malignant margins and all 27 normal margins, with only one false negative and no false positives. Similar results were seen at 10$\times$ magnification, with 98.18\% accuracy, 96.30\% sensitivity, and 100.00\% specificity at the optimal threshold of 0.232 (see Fig. \ref{fig:roc_10x}). At this magnification, most malignant samples were correctly detected, with only a small number of normal samples misclassified as malignant.

Although the optimal binary thresholds were different across magnifications, the final classification performance was the same for both TA and DL. The ROC curves showed strong separation between malignant and benign margins. For DL, the AUC values were 0.9960 at 4$\times$ and 0.9959 at 10$\times$. For TA, the AUC values were 0.9944 at 4$\times$ and 0.9938 at 10$\times$ (see Fig. \ref{fig:roc}). Based on the results, magnification primarily affects the distribution of prediction scores, but does not significantly alter classification performance. The best binary thresholds were selected via grid search along the ROC curve to maximize margin-level accuracy. Overall, TA and DL showed similar margin-level performance at both magnifications. Using a higher-resolution 10$\times$ objective did not affect classification performance at the margin-level. This shows that a lower magnification of 4$\times$ already provides sufficient tissue structure and texture information for reliable classification. It also reduces computational cost and scanning time without degrading margin-level performance in real-time use.

\begin{figure*}[t]
\centering
\includegraphics[width=\linewidth]{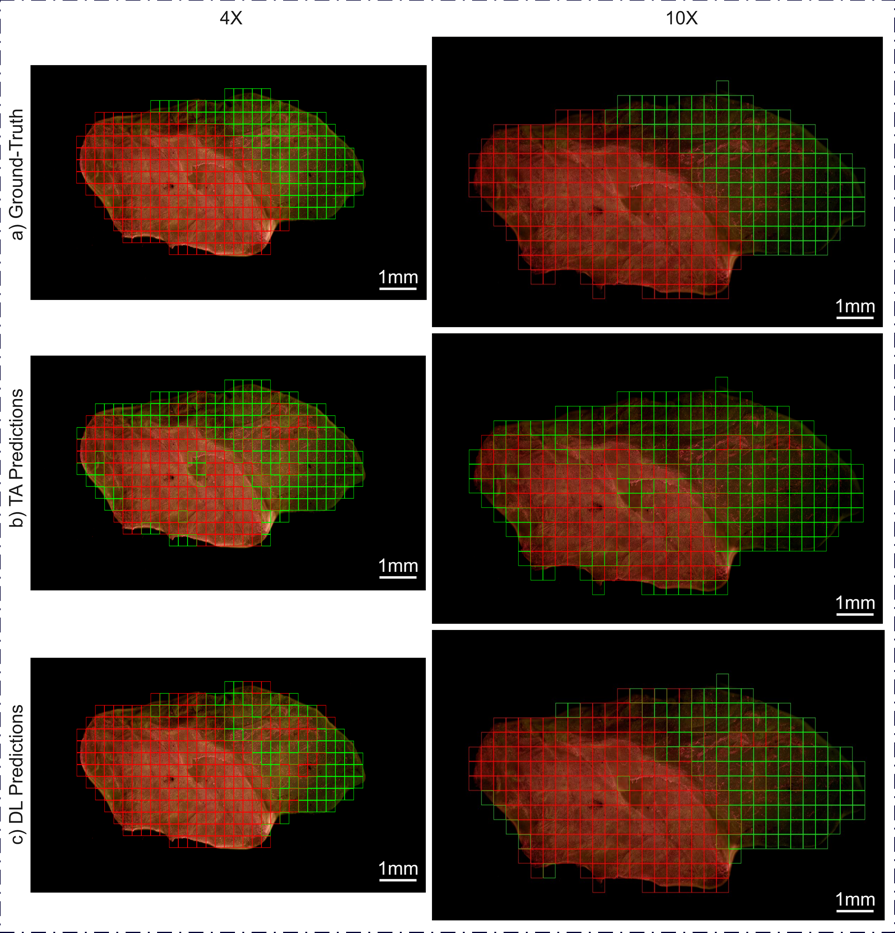} 
\caption{
Visualization of classification results at 4$\times$ (left column) and 10$\times$ (right column) magnifications. 
(a) Ground truth labels overlaid on MUSE images, (b) TA predictions, and (c) DL predictions. 
Ground truth annotations are derived from corresponding H\&E images.
}
\label{results}
\end{figure*}

\begin{figure*}[t]
    \centering
    \begin{subfigure}[b]{0.497\linewidth}
        \centering
        \includegraphics[width=\linewidth]{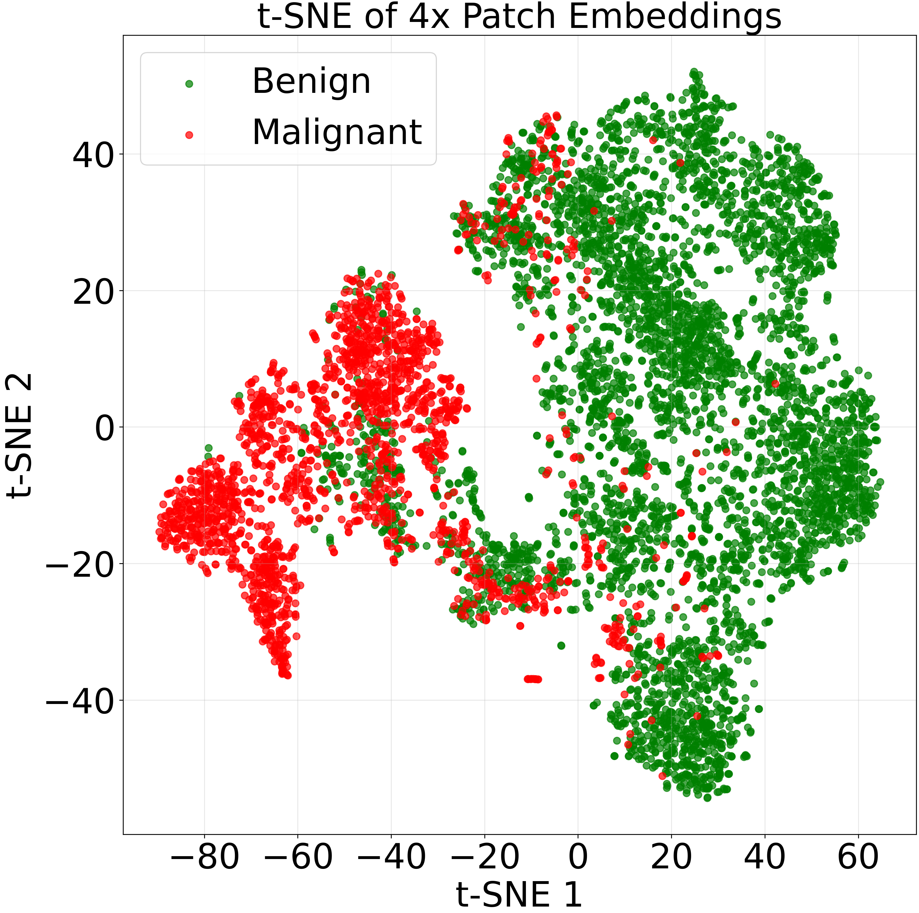}
        \caption{TSNE plot of 4$\times$}
        \label{fig:tsne_4x}
    \end{subfigure}
    \hfill
    \begin{subfigure}[b]{0.497\linewidth}
        \centering
        \includegraphics[width=\linewidth]{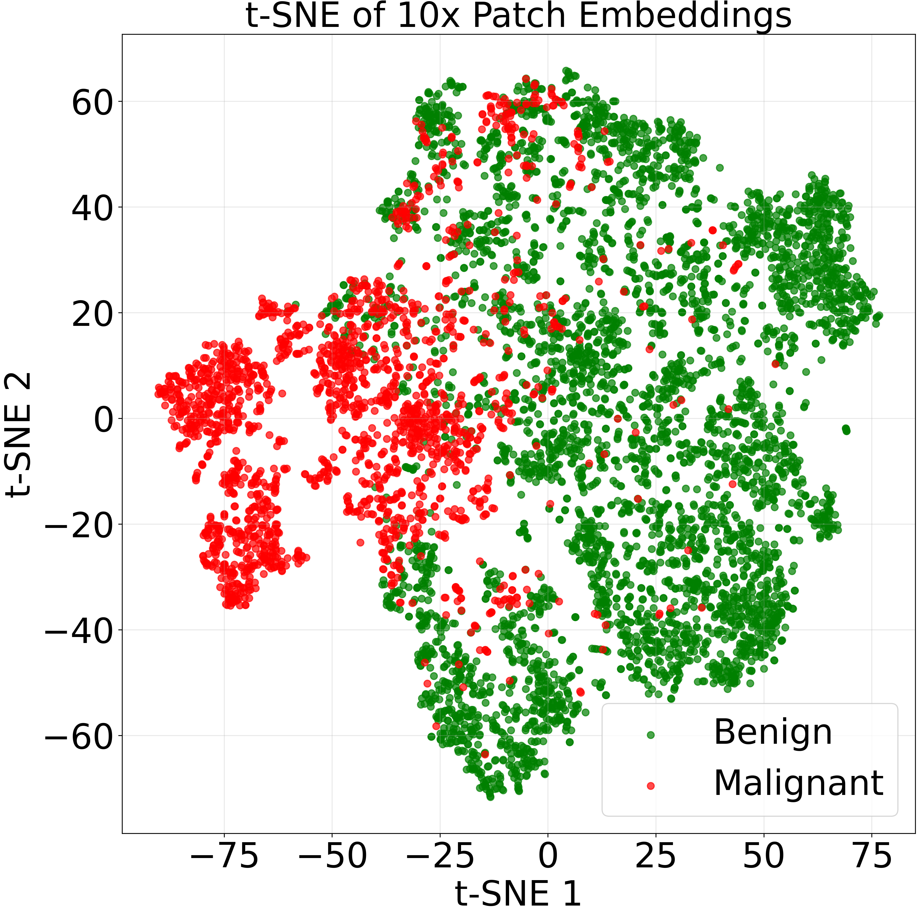}
        \caption{TSNE plot of 10$\times$}
        \label{fig:tsne_10x}
    \end{subfigure}
    \caption{TSNE plots of visualization of features across both magnifications.}
    \label{fig:tsne}
\end{figure*}

\subsection{Qualitative Results}
Fig. \ref{results} presents examples of correctly classified samples at both 4$\times$ and 10$\times$ magnifications using TA and DL methods. The figure shows patch-level prediction results for representative tissue samples, where (a) shows ground-truth labels overlaid on MUSE images, (b) shows TA prediction maps, and (c) shows DL prediction maps. The results demonstrate that the sample was correctly classified at both magnifications using both methods. 
As shown in the figure, malignant and benign regions are accurately distinguished, with most patches correctly predicted by both TA and DL models. Although a few patches exhibit minor misclassifications, the overall patch fusion results lead to correct image-level classification. These findings indicate that both TA and DL methods can effectively capture discriminative histological features across magnifications, supporting robust performance for surgical margin assessment.

To better understand the structure of the high-dimensional MUSE image patches, t-distributed Stochastic Neighbor Embedding (t-SNE) \cite{VanderMaaten2008} was used. In this method, high-dimensional data are projected into two- or three-dimensional spaces while preserving local relationships among points. t-SNE also preserves important global structures, making it effective for exploring patterns in high-dimensional data.

As depicted in Figure \ref{fig:tsne}, the t-SNE plots show feature representation for both magnifications from the ViT class token, which summarizes the global structure of each patch. Both 4$\times$ and 10$\times$ patches form well-separated clusters for benign and malignant tissue, showing that the ViT learns useful features even before the final classification layer. The 4$\times$ clusters the features slightly better, whereas the 10$\times$ feature representations are more mixed at the edges, which might be due to local variation in fine cellular details and downsampling. Some overlap is observed across both magnifications, likely reflecting the model's difficulty in distinguishing features or where benign and malignant patterns appear similar or ambiguous. Overall, most patches form clear clusters, which support the use of the patch-level framework for MUSE-level classification. Even after downsampling, the ViT still captures meaningful features.

\section{Discussion}
\subsection{Main Findings}
In this study, the effects of magnifications 4$\times$ and 10$\times$ on the classification at the margin-level of MUSE images were investigated using TA and DL.  Both methods demonstrated consistent performance, indicating that automated approaches can distinguish benign from malignant breast tissue. Specifically, TA achieves a similar general accuracy at the optimal threshold across magnifications; however, the magnification 4$\times$ shows better performance in detecting benign samples, while the magnification 10$\times$ demonstrates a greater sensitivity to identify malignant tissues. In contrast, the DL method exhibited stable, consistent performance at both magnifications. Overall, both methods achieved high margin-level accuracy, with more than 96\% accuracy at 4$\times$ and 10$\times$.

\subsection{Analysis of Misclassified Cases}
\begin{figure*}[t]
    \centering
    \includegraphics[width=0.89\linewidth]{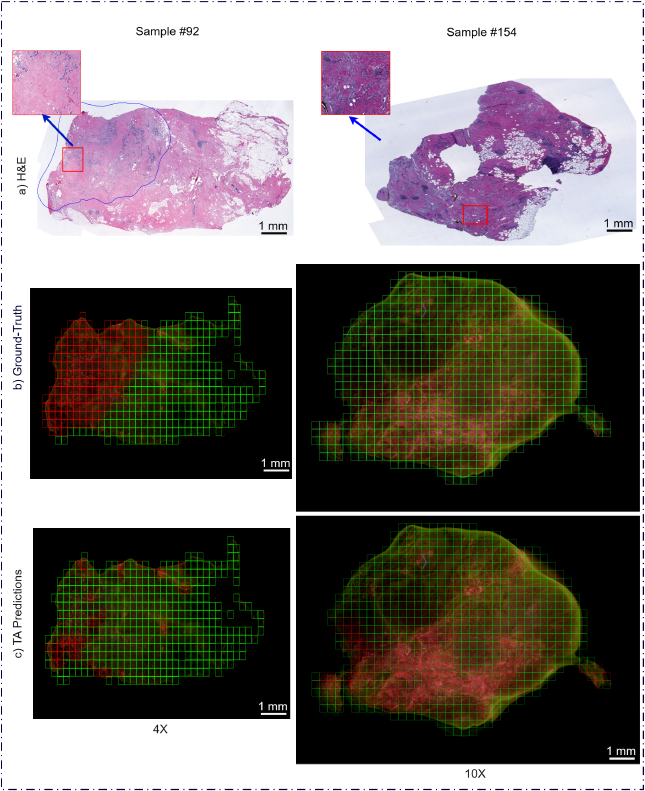}
\caption{
Examples of misclassified samples using the TA method at 4$\times$ (left column) and 10$\times$ (right column). 
The left sample is malignant but misclassified as negative, whereas the right sample is benign but misclassified as malignant. 
(a) Corresponding H\&E images for each sample, including zoomed-in patches highlighting the misclassified regions. 
(b) Ground-truth labels overlaid on the MUSE images. 
(c) TA predictions overlaid on the MUSE images.
}
    \label{fig:ta_misclassified_positive}
\end{figure*}

Fig. \mbox{\ref{fig:ta_misclassified_positive}} shows examples of TA-misclassified samples at two different magnifications. On the left, (a) shows the H\&E image of a malignant sample, (b) shows the corresponding ground-truth labels overlaid on the MUSE images, (c) shows the TA prediction map overlaid on the 4$\times$ MUSE images. In the right column, (a) shows a benign sample H\&E that was misclassified as positive (malignant), (b) shows the corresponding ground-truth labels, (c) shows the TA prediction map overlaid on the MUSE image.

According to the results in Fig. \ref{fig:ta_misclassified_positive}, misclassifications were primarily due to local texture similarities between benign and malignant tissues rather than to limitations in spatial resolution. At 4$\times$ magnification, some dark malignant regions were misinterpreted as healthy tissue at the patch level, thereby reducing the number of malignant patches and leading to an overall false-negative classification. At 10$\times$ magnification, more detailed cellular structures were visible, but some benign lobules and glandular regions were misclassified as malignant because their nuclei appeared clustered, resulting in false-positive predictions. In general, these errors were due mainly to textural similarities between benign and malignant regions rather than magnification itself. This indicates that both 4$\times$ and 10$\times$ yield a similar diagnostic performance, with most errors arising from challenging tissue patterns rather than from resolution.

\begin{figure*}[t]
    \centering
    \includegraphics[width=0.89\linewidth]{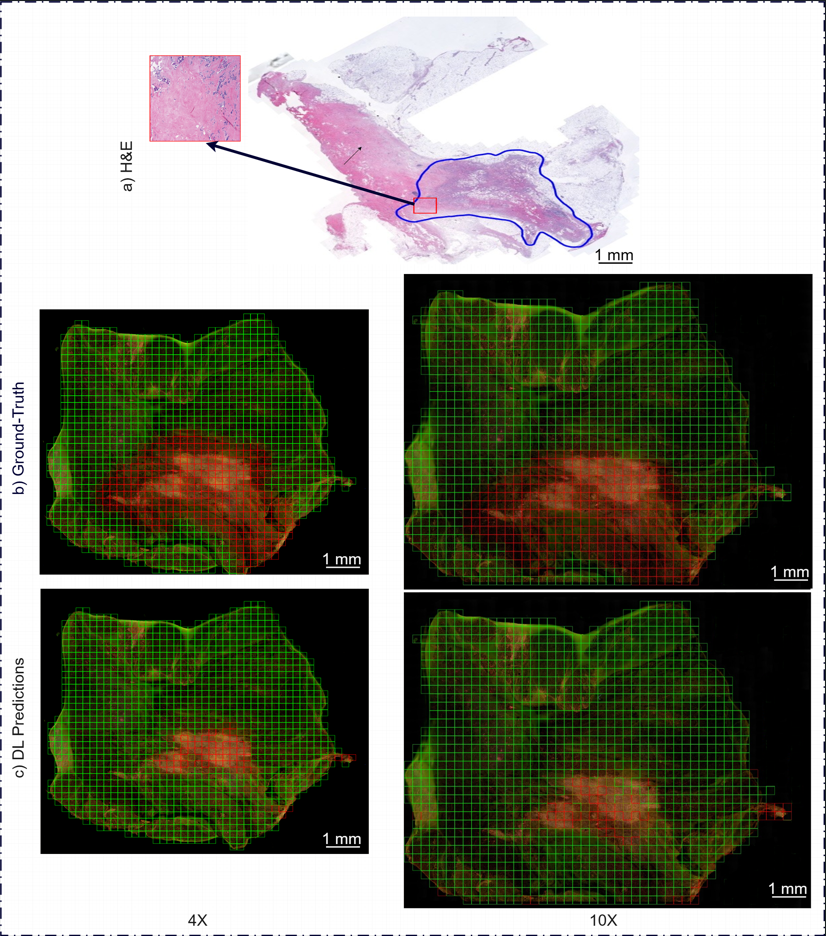}

\caption{Examples of a misclassified malignant sample using the DL method at 4$\times$ (left column) and 10$\times$ (right column). 
The same case was incorrectly predicted as benign at both magnifications. 
(a) H\&E images for each sample, including zoomed-in patches highlighting the misclassified regions. 
(b) Ground-truth labels overlaid on the MUSE images. 
(c) DL predictions overlaid on the MUSE images.
}
    \label{fig:dl_miclassified}
\end{figure*}

Fig. \ref{fig:dl_miclassified} displays the same misclassified malignant sample at both 4$\times$ and 10$\times$ magnifications by DL. (a) corresponds to the H\&E image of a malignant sample with highlighting the malignant patch in the  H\&E image. (b) shows the ground truth patches overlaid on 4$\times$ (left) and 10 $\times$ (right) magnifications. (c) displays the DL predictions overlaid for these magnifications.

As shown in the DL predictions (Fig. \ref{fig:dl_miclassified}c), for both the 4$\times$ and 10$\times$ magnifications, models correctly classified most benign patches and successfully detected the main malignant core. However, at both cases, models failed to classify several malignant patches near the tumor boundary, which resulted in false-negative regions compared with the ground truth annotations (Fig. \ref{fig:dl_miclassified}b). Importantly, the misclassified regions were very similar at both magnifications, suggesting that the errors were not primarily due to the imaging magnification. These results show that the misclassified areas contain mixed or unclear features, which the model failed to classify correctly. Because the final decision is based on majority voting, samples with too few correctly identified malignant patches were ultimately classified as false negatives.

\subsection{Limitations and Future Directions}

Objective magnification is an important design parameter for the MUSE system \cite{Kingslake2012, Frolov2018}. Although higher magnification provides finer tissue details, image resolution alone should not determine system optimization. For classification tasks, the algorithm's ability to handle different resolutions must be considered, as increased magnification does not always improve performance. In this study, the performance of a 10$\times$ objective lens was compared with that of a 4$\times$ lens using TA and DL methods. While the balance between sensitivity and specificity was affected at 10$\times$, overall accuracy was not improved, and the longer acquisition time makes it less suitable for time-sensitive intraoperative use. These findings can also inform other optical imaging systems that apply LBP for TA and ViT for DL in classification tasks \cite{Gheflati2022,Naresh2015}.

Several limitations should be acknowledged in this study. First, because our comparison is conducted at the margin level on identical specimens rather than patch-by-patch, the 4$\times$ and 10$\times$ images necessarily capture different contextual and structural information, patch sizes are scaled by magnification (2.5$\times$) to provide comparable tissue context rather than to enforce exact physical-area equivalence, and this difference reflects the magnification effect under study rather than a confound \cite{Akbarzadeh2020}. While analyzing the whole-slide images without patch-based processing would be ideal, the stitched MUSE images are exceptionally large (often exceeding 10,000 × 10,000 pixels). Passing images of this dimensionality directly through a ViT or extracting LBP features without aggressive downsampling exceeds current GPU memory limits and destroys the fine micro-textural details required for diagnosis. Thus, patch-based processing is a computational necessity.

Second, the 10$\times$ objective lens has a limited depth of field, and some regions of the MUSE images may be slightly out of focus, potentially affecting classification accuracy. Third, the dataset was imbalanced at the patch level, with more than 75\% benign patches and less than 25\% malignant patches at both magnifications. This scenario might affect the margin-level decision. In the case of small malignant areas, where majority voting may allow the larger number of benign patches to dominate, this can lead to false-negative margin predictions. 

Future work should consider adaptive patch-sizing methods that better capture tissue context across different magnifications. In addition, the models were not externally validated on an independent dataset, which will be necessary to confirm generalizability. Automated focusing techniques might be worth applying to improve image quality, especially at higher magnifications. In addition, a weighted majority voting method can be explored to improve MUSE's margin-level classification by assigning greater weight to suspicious patches. Collecting more malignant samples will help balance the dataset and improve the model's robustness and generalizability.

\section{Conclusion}
This study investigated how 4$\times$ and 10$\times$ magnifications affect margin-level classification of MUSE images using both TA and DL methods, employing a patch-level framework. For TA, the red channel of each patch was preprocessed, LBP features were extracted, and an SVM predicted the patch-level class. For DL, patch embeddings were processed through a ViT with an MLP head to get patch-level predictions. In both methods, patch predictions were combined and compared to a binary threshold to predict the final MUSE image prediction.  Quantitative results showed that DL and TA performed consistently across both magnifications, though TA was slightly less accurate. Overall, 10$\times$ magnification did not improve margin-level accuracy compared to 4$\times$. Since 10$\times$ takes longer to acquire and requires more computation, 4$\times$ is a practical and efficient choice for intraoperative margin assessment. Overall, MUSE has the potential to require less time than traditional histopathology methods, such as frozen-section analysis, and to provide diagnostic results using DL, thereby reducing the need for manual image interpretation by pathologists. These findings can help develop more accurate diagnostic models, support careful margin inspection during breast-conserving surgery, and improve patient outcomes.

\begin{backmatter}
\bmsection{Funding}
National Institute of Biomedical Imaging and Bioengineering of the National Institutes of Health under Award Number R01EB033806. The content is solely the responsibility of the authors and does not necessarily represent the official views of the National Institutes of Health.

\bmsection{Acknowledgment}
The authors thank the Medical College of Wisconsin Tissue Bank for providing de-identified human breast tissue specimens.  

\bmsection{Disclosures}
The authors have no conflicts of interest to claim. 

\bmsection{Data Availability Statement}
Data underlying the results presented in this paper are not publicly available at this time but may be obtained from the authors upon reasonable request.

\end{backmatter}

\end{document}